\documentclass[11pt]{article}

\usepackage[final]{acl}

\usepackage{times}
\usepackage{latexsym}

\usepackage[T1]{fontenc}

\usepackage[utf8]{inputenc}

\usepackage{microtype}

\usepackage{inconsolata}

\usepackage{graphicx}

\usepackage{times}
\usepackage{latexsym}
\usepackage[T1]{fontenc}
\usepackage[utf8]{inputenc}
\usepackage{microtype}
\usepackage{inconsolata}
\usepackage{graphicx}

\usepackage{amsmath, amssymb}
\usepackage{booktabs}
\usepackage{stmaryrd}
\usepackage{array}
\usepackage{multirow}
\usepackage{algorithm}
\usepackage{algpseudocode}
\usepackage{mathabx}

\usepackage{pifont}
\newcommand{\cmark}{\ding{51}}
\newcommand{\xmark}{\ding{55}}

\usepackage{bm}

\title{Caliper: Probing Lexical Anchors versus Causal Structure in LLMs}

\author{Zhenyu Yu \\
  Fudan University \\
  \texttt{yuzhenyuyxl@foxmail.com} \\\And
  Shuigeng Zhou \\
  Fudan University \\
  \texttt{sgzhou@fudan.edu.cn} \\}

\begin{document}
\maketitle
\begin{abstract}
Large language models reach 50 to 70\% accuracy on causal reasoning benchmarks such as CLadder, but it is unclear whether this reflects structural reasoning or lexical pattern matching. We introduce \textbf{Caliper}, a controlled perturbation that replaces semantic variable names with placeholder tokens while preserving the causal graph and probabilistic specification of each question. Across nine instruction-tuned LLMs from 3.8B to 671B and three causal reasoning benchmarks, lexical anonymization yields robust accuracy drops of $+7.6$, $+27.0$, and $+11.1$ pp on a local 3.8B--14B set, rising to $+29.6$ and $+18.0$ pp on CRASS and e-CARE across nine frontier models spanning the 2024--2026 generations. Of 40 engaged model-by-benchmark cells, 39 show a positive gap, and the gap collapses by $17\times$ on CLadder's pseudoword subset. Structured scaffolding and few-shot in-context learning each narrow the gap, but mainly by lowering $P_0$ accuracy on smaller models rather than recovering $P_1$. Current instruction-tuned LLMs, evaluated zero-shot, show little evidence of structural causal reasoning once lexical anchors are removed.
\end{abstract}

\begin{figure*}[t]
    \centering
    \includegraphics[width=1.0\linewidth]{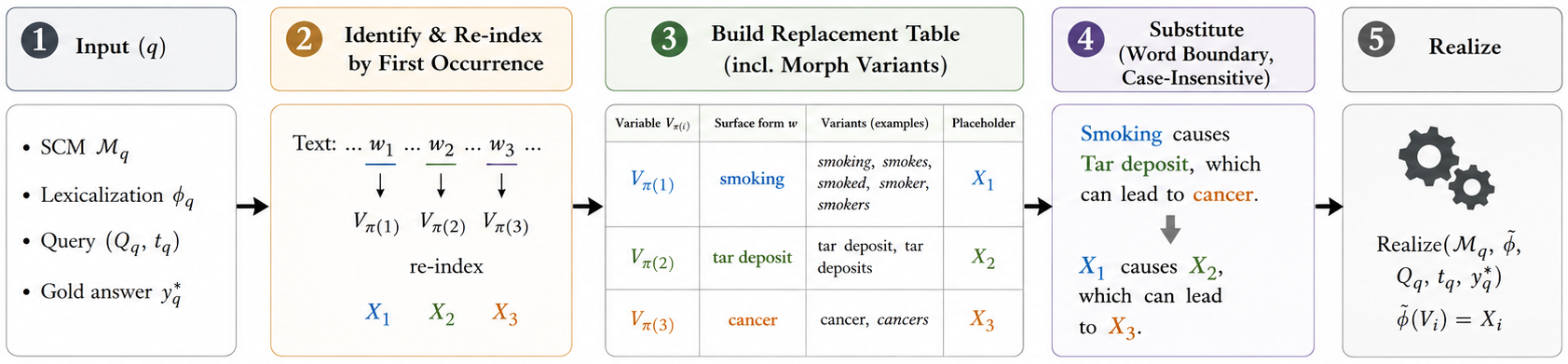}
    \caption{Overview of the Caliper perturbation. Each variable surface form in a causal-QA item is replaced with a placeholder symbol $X_i$, while the causal graph, joint distribution, query, and gold answer are preserved.}
    \label{fig:caliper-overview}
\end{figure*}

\section{Introduction} \label{sec:intro}

Whether large language models reason \emph{causally} or pattern match on the surface form of causal language remains contested~\citep{liu2025causalsurvey, shrestha2025parrotsprophets, miliani2025explica}. The CLadder benchmark of \citet{jin2023cladder} reports GPT-4 at 62\% accuracy on a Pearl ladder battery while LLaMa-1 reaches only 44\%. The size gap admits two readings. Larger models may have acquired an emergent causal inference ability~\citep{yu2025causaleval, zhou2024causalbench, feng2025reliability}. Alternatively, they may simply hold richer lexical priors over common English variable names such as \emph{smoking} and \emph{cancer}, and those priors substitute for inference on benchmark questions that happen to use the same surface vocabulary~\citep{chi2024unveiling, chen2025contamination, menta2025memorization}. On the standard benchmark the two hypotheses are observationally indistinguishable, since every question carries both its causal structure and its everyday English wording~\citep{dehghanighobadi2025counterfactually}.

To disentangle these two readings, we introduce \textbf{Caliper}, a controlled perturbation that preserves the causal graph and the probabilistic specification of each question while replacing the semantic variable names with placeholder tokens. \textit{Smoking} becomes $X_1$ and \textit{cancer} becomes $X_2$, and the rest of the question is left exactly as it was. Any accuracy drop under this perturbation cannot be attributed to a change in the underlying causal task and must reflect a loss of surface lexical content. The contested question becomes a falsifiable measurement. A model with structural causal reasoning ability should be invariant under Caliper, whereas a model that relies on lexical priors should not.

We apply this measurement to fourteen instruction-tuned LLMs on three causal reasoning benchmarks, CLadder, CRASS, and e-CARE. The local set spans 3.8B--14B (Phi-3-mini, Qwen2.5-7B, Mistral-7B-v0.3, Llama-3.1-8B, AWQ-quantized Qwen2.5-14B). The frontier set spans nine open-weight models across the Llama, Qwen, DeepSeek, and Hermes-3 families. The analysis dissects the perturbation effect across Pearl rungs, fine-grained query types, story domains, and the pre-existing nonsensical subset of CLadder, and tests whether three standard prompting interventions, a structured scaffolding prompt, few-shot in-context learning, and the CausalCoT prompt of \citet{jin2023cladder}, can recover the accuracy that Caliper removes.

\paragraph{Contributions.}
\begin{itemize}
  \item \textbf{Caliper perturbation.} A structure-preserving lexical perturbation grounded in CLadder's structural causal model (SCM) metadata, with a part-of-speech aware extension to multiple-choice questions (MCQ) benchmarks without SCM annotation.
  \item \textbf{Lexical mediation of causal accuracy.} The Caliper gap localises to interventional and counterfactual queries, concentrates on ATE items and high-prior story domains, and vanishes on the pre-existing nonsensical subset, identifying surface lexical content as the mediator of benchmark performance.
  \item \textbf{Persistence across scale and prompting.} The gap replicates on CRASS, e-CARE, and four frontier models from 70B to 671B, and resists structured scaffolding, few-shot in-context learning, and CausalCoT.
\end{itemize}

\section{Related Work} \label{sec:related}

\paragraph{Benchmarking causal reasoning in LLMs.} CLadder \citep{jin2023cladder} is a dataset of roughly 10{,}000 yes/no questions stratified by the three rungs of Pearl's causal hierarchy~\citep{pearl2009causality} and grounded in symbolic structural causal models. CRASS \citep{frohberg2022crass} and e-CARE \citep{du2022ecare} provide multiple-choice counterfactual and commonsense causal reasoning tasks. Beyond these question-answering benchmarks, recent work probes adjacent capabilities such as causal discovery \citep{zhou2024causalbench, feng2025reliability} and explicit-connective causal reasoning \citep{miliani2025explica}, and surveys methodological efforts to strengthen LLM causal performance \citep{liu2025causalsurvey, yu2025causaleval}. Across these efforts, large size gaps between GPT-4 and smaller models have been read both as evidence for emergent causal inference \citep{kiciman2023causalbench} and, in tension, as evidence for sophisticated pattern matching \citep{zecevic2023causalparrots}. Our work provides a direct probe that disentangles the two readings.

\paragraph{Surface perturbation probes.} Surface perturbation is a long-standing probe of model robustness \citep{ribeiro2020checklist}, and spurious surface-form-to-label correlations have been shown to drive performance in NLI \citep{mccoy2019right} and commonsense benchmarks \citep{elazar2021back}. Memorization of pretraining text has been quantified by \citet{carlini2022quantifying} and exploited in evaluation by \citet{magar2022data}, with recent work refining the measurement of memorization through probabilistic extraction \citep{hayes2025probabilistic} and model-attribution analysis \citep{menta2025memorization}. A parallel line of work \citep{chen2025contamination} surveys data-contamination effects on LLM evaluation and motivates moving from static to dynamic benchmarks. Caliper extends this line of work to causal reasoning, where the structural task is fixed by an SCM and surface lexicalization can be cleanly factored out.

\paragraph{Causal parrots.} The parroting hypothesis, articulated qualitatively by \citet{zecevic2023causalparrots}, posits that LLMs may answer causal questions by parroting memorized correlations rather than reasoning over a causal model. Recent work has quantified this concern through several routes. \citet{chi2024unveiling} construct the CausalProbe-2024 benchmark from post-cutoff news articles to factor out training-data overlap, concluding that LLMs are limited to level-1 causal reasoning. \citet{shrestha2025parrotsprophets} revisit the claim on CLadder and test structured prompting as a mitigation, and \citet{dehghanighobadi2025counterfactually} show that LLM-generated counterfactual explanations frequently disagree with the model's own predictions. Caliper takes a complementary route by leaving the benchmark fixed and varying only the lexical surface form. It provides quantitative cross-architecture evidence on fourteen LLMs across three benchmarks, with the effect localised to specific query types and story domains.

\section{The Caliper Framework} \label{sec:method}

\subsection{Preliminaries} \label{sec:prelim}

\paragraph{Structural causal models.} We follow the standard definition of an SCM as a tuple $\mathcal{M}\!=\!\langle V, U, F, P(U)\rangle$ over endogenous variables $V$, exogenous noise $U$, structural equations $F$, and a noise distribution $P(U)$ \citep{pearl2009causality}. The induced DAG $\mathcal{G}(\mathcal{M})$ has an edge $V_j\!\to\!V_i$ whenever $V_j\!\in\!\mathrm{Pa}(V_i)$.

\paragraph{Pearl ladder.} Pearl's causal hierarchy partitions causal queries into three rungs:
\begin{align}
& \text{Rung 1 (Obs.):}\quad P(Y\!\mid\!X\!=\!x), \label{eq:rung1}\\
& \text{Rung 2 (Int.):}\quad P(Y\!\mid\!do(X\!=\!x)), \label{eq:rung2}\\
& \text{Rung 3 (CF):}\quad   P(Y_{X=x}\!\mid\!X\!=\!x', Y\!=\!y'),
\label{eq:rung3}
\end{align}
where $do(\cdot)$ denotes a structural intervention that removes the equation for $X$ and fixes $X\!=\!x$, and the counterfactual in Eq. \eqref{eq:rung3} is well-defined for deterministic-functional SCMs.

\paragraph{SCM-grounded causal QA.} We assume a causal-QA item $q$ specified as a tuple $\langle\mathcal{M}_q, \phi_q, Q_q, t_q, y^\star_q\rangle$ consisting of an SCM $\mathcal{M}_q$, a \emph{lexicalization map} $\phi_q : V\!\to\!\mathcal{L}$ that assigns each SCM variable $V_i$ a surface English realization $\phi_q(V_i)$, a query $Q_q$ at one of the three rungs together with a query-type label $t_q$, and a gold answer $y^\star_q$. Caliper is defined for any benchmark with this structure, and in our experiments the running example is CLadder \citep{jin2023cladder}, whose \texttt{variable\_mapping} field provides $\phi_q$ and whose generator instantiates $\mathrm{Realize}$ below. A typical lexicalization on the smoking--cancer story gives $\phi_q(V_1)\!=\!\emph{smoking}$, $\phi_q(V_2)\!=\!\emph{tar deposit}$, $\phi_q(V_3)\!=\!\emph{cancer}$. 

\subsection{Definition and Invariance} \label{sec:perturbation}

\paragraph{Definition.} Let $\tilde{\phi}$ be the \emph{anonymous lexicalization map} that sends every SCM variable $V_i$ to the placeholder symbol $X_i$:
\begin{equation}
\tilde{\phi}(V_i) \;=\; X_i, \qquad i = 1, \ldots, n.
\label{eq:anon-map}
\end{equation}
The Caliper transformation $T$ takes a question $q\!=\!\langle\mathcal{M}_q, \phi_q, Q_q, t_q, y^\star_q\rangle$ and returns
\begin{equation}
T(q) \;=\; \mathrm{Realize}\bigl(\mathcal{M}_q, \tilde{\phi},
                                 Q_q, t_q, y^\star_q\bigr),
\label{eq:caliper-def}
\end{equation}
where $\mathrm{Realize}$ is the benchmark's deterministic surface-form generator. The pipeline is illustrated in Figure~\ref{fig:caliper-overview}, and the explicit implementation is given in Algorithm~\ref{alg:caliper}. We write $P_0(q)\!=\!q$ for the original question and $P_1(q)\!=\!T(q)$ for its lexically anonymized counterpart.

\paragraph{Invariance properties.} By construction, $T$ preserves four properties of $q$:
\begin{align}
\mathcal{G}\bigl(\mathcal{M}_{T(q)}\bigr) &= \mathcal{G}(\mathcal{M}_q),
\label{eq:inv-graph}\\
P_{T(q)}(V, U) &= P_q(V, U),
\label{eq:inv-prob}\\
\bigl(Q_{T(q)}, t_{T(q)}\bigr) &= \bigl(Q_q, t_q\bigr),
\label{eq:inv-query}\\
y^\star_{T(q)} &= y^\star_q.
\label{eq:inv-gold}
\end{align}

\paragraph{Identifiability implication.} For any oracle reasoner $R$ that operates only on the SCM $\mathcal{M}_q$, the query $Q_q$, and the structural equations $F_q$, the answer on $T(q)$ is necessarily identical to the answer on $q$:
\begin{equation}
R(q) \;=\; R(T(q)) \quad \forall q.
\label{eq:oracle}
\end{equation}
Our experimental design is grounded in Eq.~\eqref{eq:oracle}. For a model $M$, any deviation $M(T(q))\!\neq\!M(q)$ on a fixed-rung subset cannot be attributed to a change in the underlying causal task and must reflect $M$'s sensitivity to the surface realization $\phi_q$. The perturbation is therefore a structure-preserving ablation rather than noise injection. The conditional probabilities, the do-operator semantics, and the SCM graph are all present in $T(q)$, so a reasoner that operates over them can still produce the correct answer. 

\subsection{Instantiation} \label{sec:algo}

\paragraph{Procedure.} 
We restrict the replacement table to variables that appear in the question text and re-index them by first occurrence, so that $X_1, X_2, \ldots$ follow reading order. Each surface form, together with its plural and morphological variants, is then mapped to its placeholder token. Substitution is performed with word-boundary matching and is case-insensitive, so \emph{Smoking}, \emph{smoking}, and \emph{SMOKING} are all replaced by the same $X_i$. Replacement keys are sorted by descending length so that multiword realizations such as \emph{tar deposit} are matched before any of their substrings.

\begin{algorithm}[t]
\caption{Caliper perturbation $T$}
\label{alg:caliper}
\begin{algorithmic}[1]
\Require Question $q$ with SCM $\mathcal{M}_q$, lexicalization $\phi_q$, gold $y^\star_q$
\Ensure Anonymized question $T(q)$
\State $s \gets \mathrm{Text}(q)$; $R \gets \{\,\}$
\State Re-index variables in $V_q$ appearing in $s$ as $V_{\pi(1)}, \ldots, V_{\pi(m)}$ by first occurrence
\For{$i = 1$ \textbf{to} $m$, $w \gets \phi_q(V_{\pi(i)})$}
  \State $R[w] \gets X_i$;\, $R[w'] \gets X_i$ \textbf{for} $w' \in \mathrm{MorphVariants}(w)$
\EndFor
\For{each key $k$ in $R$, sorted by descending length}
  \State $s \gets \mathrm{WordBoundaryReplace}(s, k, R[k])$ \Comment{case-insensitive}
\EndFor
\State \Return $\mathrm{Realize}(s, \mathcal{M}_q, Q_q, t_q, y^\star_q)$
\end{algorithmic}
\end{algorithm}

\paragraph{Morphological variants.} The set $\mathrm{MorphVariants}(w)$ covers regular plurals, common irregular forms (e.g., \emph{wife/wives}, \emph{child/children}, \emph{foot/feet}), and the \mbox{-y/-ies}, \mbox{-f/-ves}, and \mbox{-fe/-ves} rules. Verb-derived gerunds in the lexicalization map are captured automatically.

\paragraph{Scope of replacement.} Only variable surface forms listed by $\phi_q$ are anonymized. Causal connectives, probability terms, numerical values, and function words remain untouched. Using the lexicalization map rather than regex matching avoids the false positives that arise when variable names overlap with English function words. Exact word lists for our benchmark instantiations see Appendix~\ref{app:prompts}.

\paragraph{Adaptation to benchmarks without SCM annotation.} For causal-QA benchmarks that do not provide $\phi_q$, we use a part-of-speech aware variant of Caliper. Each item is part-of-speech tagged, every NOUN and PROPN token is mapped to a placeholder symbol $X_1, X_2, \ldots$ by first occurrence, and the resulting mapping is applied consistently across the premise, the question, and every option. Verbs, adverbs, prepositions, valence adjectives, causal connectives, modals, and function words are left untouched, so the structural relation between the premise and the options is preserved.

\begin{table*}[t]
\centering
\small
\setlength{\tabcolsep}{12pt}
\begin{tabular}{l ccc ccc ccc}
\toprule
 & \multicolumn{3}{c}{\textbf{Obs}} & \multicolumn{3}{c}{\textbf{Int}} & \multicolumn{3}{c}{\textbf{CF}}\\
\cmidrule(lr){2-4}\cmidrule(lr){5-7}\cmidrule(lr){8-10}
\textbf{Model} & $\bm{P_0}$ & $\bm{P_1}$ & $\bm{\Delta}$ & $\bm{P_0}$ & $\bm{P_1}$ & $\bm{\Delta}$ & $\bm{P_0}$ & $\bm{P_1}$ & $\bm{\Delta}$ \\
\midrule
Phi-3-mini & 49.4 & 52.8 & $-$3.4 & 64.8 & 58.6 & $+$6.2 & 53.8 & 51.6 & $+$2.2\\
Qwen2.5-7B & 48.8 & 48.8 & $+$0.0 & 61.8 & 52.0 & $+$9.8 & 56.8 & 55.8 & $+$1.0\\
Mistral-7B & 50.4 & 49.8 & $+$0.6 & 58.4 & 51.2 & $+$7.2 & 55.6 & 54.4 & $+$1.2\\
Llama-3.1-8B & 50.2 & 48.6 & $+$1.6 & 59.8 & 53.6 & $+$6.2 & 53.8 & 50.6 & $+$3.2\\
Qwen2.5-14B & 55.0 & 56.0 & $-$1.0 & 64.0 & 55.4 & $+$8.6 & 60.2 & 54.4 & $+$5.8\\
\midrule
\textbf{Mean} & 50.8 & 51.2 & \textbf{$-$0.4} & 61.8 & 54.2 & \textbf{$+$7.6} & 56.0 & 53.4 & \textbf{$+$2.7}\\
\bottomrule
\end{tabular}
\caption{CLadder commonsensical accuracy across three rungs under original ($P_0$) and anonymized ($P_1$) prompts; $\Delta = P_0 - P_1$ (pp).}
\label{tab:main}
\end{table*}

\section{Experiments} \label{sec:exp}

\subsection{Experimental Setup} \label{sec:setup}

\paragraph{Datasets.} The main analysis uses CLadder \citep{jin2023cladder}, from which we draw a stratified sample of 1{,}500 questions (500 per Pearl rung) from the commonsensical subset. The commonsensical subset uses common English variable names such as \emph{smoking} and \emph{cancer}. The nonsensical subset uses pseudoword variables and is held out for the falsifiability test. For cross-benchmark replication we additionally use the full CRASS test set \citep{frohberg2022crass} of 274 four-way counterfactual MCQ items and a stratified 1{,}000-item subsample of the e-CARE \citep{du2022ecare} validation split.

\paragraph{Models.} We evaluate fourteen instruction-tuned LLMs in two sets. The local set comprises five LLMs from 3.8B to 14B parameters: Phi-3-mini-4k-instruct \citep{abdin2024phi3}, Mistral-7B-Instruct-v0.3 \citep{jiang2023mistral}, Llama-3.1-8B-Instruct \citep{grattafiori2024llama3}, Qwen2.5-7B-Instruct, and the AWQ-quantized Qwen2.5-14B-Instruct \citep{qwen2025qwen25}. The frontier set comprises nine open-weight models grouped by family: three Llama models (Llama-3.3-70B-Instruct, Llama-4-Scout, Llama-4-Maverick \citep{grattafiori2024llama3}), two Qwen models (Qwen2.5-72B-Instruct, Qwen3-Max \citep{qwen2025qwen25}), three DeepSeek models (DeepSeek-V3, DeepSeek-V3.1, DeepSeek-V3.2 \citep{deepseekai2024deepseekv3}), and Hermes-3-Llama-3.1-405B \citep{nous2024hermes3}.

\paragraph{Evaluation Metrics.} We report per-rung accuracy on $P_0$ and $P_1$ and define the per-rung lexical gap as
\begin{equation}
\Delta_r(M) \;=\; \mathrm{acc}_r(M, P_0) - \mathrm{acc}_r(M, P_1),
\label{eq:gap}
\end{equation}
with cross-model mean $\bar{\Delta}_r \!=\! |\mathcal{M}|^{-1}\!\sum_{M\in\mathcal{M}} \Delta_r(M)$. Per-item confidence is the larger of the two normalised first-token probabilities on $\{\text{yes},\text{no}\}$. We test $H_0\!:\bar{\Delta}_r\!=\!0$ against $H_1\!:\bar{\Delta}_r\!>\!0$ on $\{\Delta_r(M)\}$ with the one-sided exact sign test, and report a 95\% bootstrap percentile confidence interval (CI) \citep{efron1993introduction} on $\bar{\Delta}_r$ with $B\!=\!1{,}000$ resamples.

\paragraph{Implementation Details.} Local inference uses HuggingFace Transformers with greedy decoding and fp16 weights on a single NVIDIA RTX 4090D GPU. Frontier inference uses a system message constraining the model to emit only the answer token. On CLadder we score \emph{Yes}/\emph{No} by first-token logprob; on CRASS and e-CARE we score each option by its mean per-token log-probability \citep{brown2020gpt3} and predict the argmax. Predictions that emit no parseable answer token are marked \emph{invalid}, with rates of 0.0--0.5\% under Direct, at most 12\% under Scaffold and Few-shot in-context learning (ICL), and higher under CausalCoT.

\subsection{Comparison across Causal Rungs} \label{sec:main}

Per-model accuracy on $P_0$ and $P_1$ across the three rungs is reported in Table~\ref{tab:main} and visualised in Appendix Figure~\ref{fig:rung}. The cross-model mean interventional gap is $+7.6$\,pp and is positive in all five architectures, with a 95\% bootstrap CI of $[+6.4,+8.8]$\,pp. The one-sided exact sign test rejects the null of zero gap at $p \approx 0.031$, corresponding to all five per-model gaps being positive. The observational gap is $-0.4$\,pp with a CI of $[-2.1,+0.9]$\,pp. This clean null rules out a generic task-difficulty confound that would degrade all rungs equally. The counterfactual gap is positive but smaller, at $+2.7$\,pp with a CI of $[+1.3,+4.2]$\,pp. 
This pattern runs against the conventional Pearl-ladder difficulty ordering, in which counterfactual reasoning is typically described as the most demanding rung, yet under our lexical perturbation it is \emph{less} affected than interventional reasoning. We attribute this to the structure of CLadder's counterfactual questions, which already involve non-lexicalized conditioning machinery (a factual world and a counterfactual world) that contributes most of the answer signal independently of the variable names.

\begin{figure}[t]
\centering
\includegraphics[width=\linewidth]{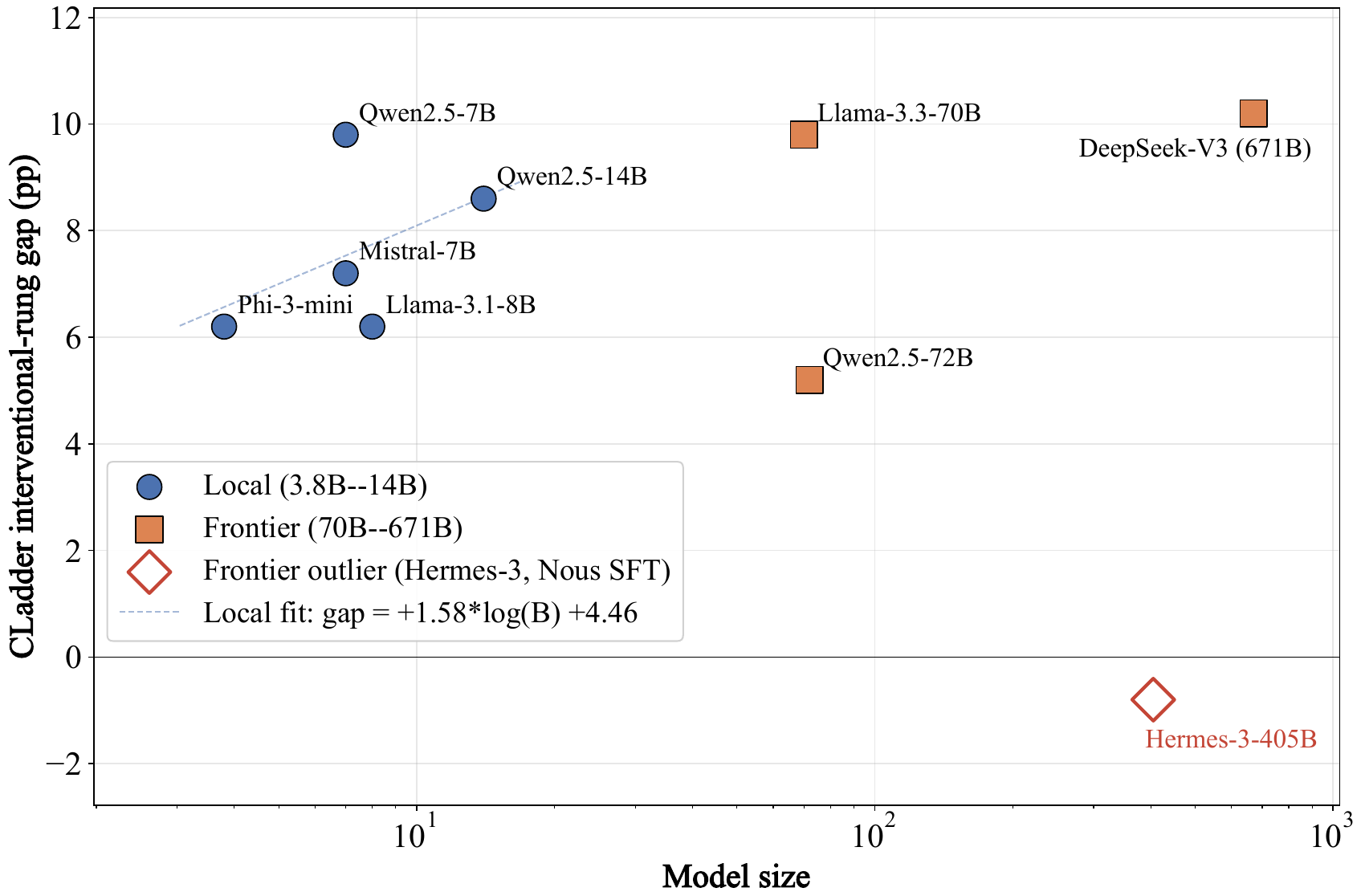}
\caption{CLadder interventional gap by model size (log scale). Local set (blue circles) fit by a log-linear trend; frontier set (orange squares) overlaid. Hermes-3-405B (red diamond) is the only reversal and is CLadder-specific.}
\label{fig:scale}
\end{figure}

\subsection{Query-Type Analysis}
\label{sec:qtype}

The interventional rung is decomposed into its three query types in Table~\ref{tab:querytype} and Appendix Figure~\ref{fig:qtype}, with full per-model per-query-type accuracies given in Appendix~\ref{app:full-qtype}\footnote{Query-type labels follow the CLadder \texttt{query\_type} field; see Appendix Table~\ref{tab:querytype-glossary} for the full-term mapping.}. The gap is heavily concentrated in \emph{ATE} queries at $+10.7 \pm 4.3$\,pp, with the smaller interventional types contributing $+7.7 \pm 4.2$\,pp for \emph{collider\_bias} and $+4.8 \pm 5.2$\,pp for \emph{backadj}. All three observational sub-types show gaps statistically indistinguishable from zero. ATE is the canonical interventional question of the form ``does treatment $X$ cause outcome $Y$''. It corresponds most directly to the form of causal claim a model could memorize from pretraining text, for example ``smoking causes cancer''. The concentration of the gap on ATE queries is consistent with the lexical memorization account.

\begin{table}[t]
\centering
\small
\setlength{\tabcolsep}{3pt}
\begin{tabular}{ll ccc}
\toprule
\textbf{Rung} & \textbf{Query type} & \textbf{$\bm{P_0}$ (\%)} & \textbf{$\bm{P_1}$ (\%)} & \textbf{$\bm{\Delta}$ (pp)} \\
\midrule
Obs & marginal & 50.5 & 50.4 & $+$0.2\,$\pm$\,2.0 \\
 & correlation & 50.5 & 51.2 & $-$0.7\,$\pm$\,2.8 \\
 & exp\_away & 54.5 & 58.6 & $-$4.1\,$\pm$\,8.0 \\
\midrule
Int & \textbf{ate} & 72.0 & 61.3 & \textbf{$+$10.7\,$\pm$\,4.3} \\
 & collider\_bias & 58.5 & 50.8 & $+$7.7\,$\pm$\,4.2 \\
 & backadj & 53.1 & 48.3 & $+$4.8\,$\pm$\,5.2 \\
\midrule
CF & nie & 52.2 & 49.0 & $+$3.2\,$\pm$\,5.4 \\
 & nde & 54.3 & 51.3 & $+$3.0\,$\pm$\,6.8 \\
 & det-counterfactual & 60.6 & 57.7 & $+$2.8\,$\pm$\,1.8 \\
 & ett & 54.2 & 52.0 & $+$2.1\,$\pm$\,1.1 \\
\bottomrule
\end{tabular}
\caption{Cross-model mean gap $\Delta$ ($\pm$ standard deviation) by Pearl-ladder query type, $n=5$ models.}
\label{tab:querytype}
\end{table}

\begin{figure*}[t]
\centering
\includegraphics[width=\linewidth]{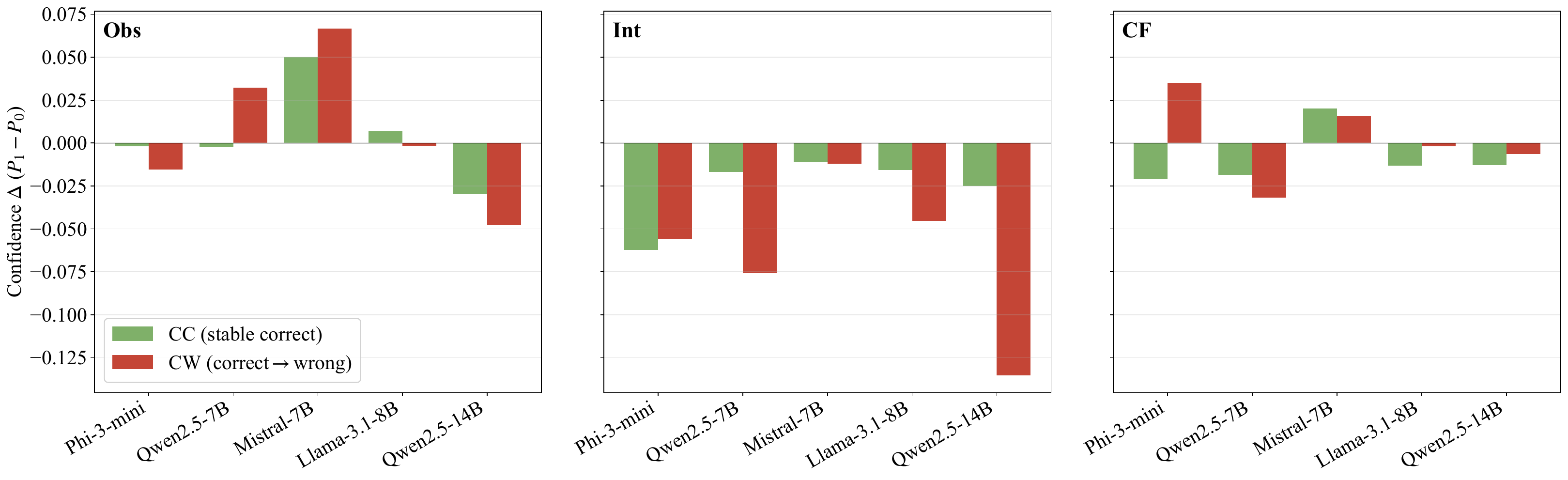}
\caption{Confidence shift ($P_1 - P_0$) on stable-correct items (CC, green) and flipped items (CW, red) by rung. On the interventional panel the CW drop is consistently larger than the CC drop.}
\label{fig:conf}
\end{figure*}

\subsection{Scaling Behaviour}
\label{sec:scale}

Within the local set we observe a non-monotonic scaling pattern on the ATE-specific gap. Qwen2.5-14B shows the smallest ATE gap at $+3.1$ pp, the 7B--8B models cluster between $+13$ and $+14$ pp, and Phi-3-mini at 3.8B sits below them at $+8.6$ pp, plausibly because its training mixes a synthetic textbook corpus \citep{abdin2024phi3} that shifts the memorization profile. 
The interventional-rung mean gap is plotted against model size in Figure~\ref{fig:scale} across all 14 evaluated models. The nine frontier models span 70B to 671B (total parameters) and two release generations (2024 and 2025--2026). Their per-model gaps range from $-0.8$ to $+10.2$ pp, with a frontier mean of $+6.7$ pp, within 1 pp of the local mean of $+7.6$ pp. Scaling from 14B to 671B does not close the gap; the 2025--2026 generation (Llama-4, Qwen3-Max, DeepSeek-V3.1/V3.2) shows gaps of $+5.0$ to $+9.8$ pp, fully consistent with the 2024 generation. 
The single direction reversal on Hermes-3-405B is CLadder-specific (CRASS $+33.2$, e-CARE $+18.8$ pp; Table~\ref{tab:cross-benchmark}). This model is Nous Research's non-standard SFT/DPO tuning of Llama-3.1-405B-base, distinct from the first-party chat tunings of the other frontier models, and its CLadder reversal suggests that the tuning recipe interacts specifically with the yes/no template rather than reflecting structural robustness to lexical perturbation.

\subsection{Confidence Analysis}
\label{sec:conf}

We split items into two outcome classes by joint correctness on $P_0$ and $P_1$. The class CC denotes items the model gets correct on both. The class CW denotes items the model gets correct on $P_0$ but wrong on $P_1$, that is, items flipped by the perturbation. If $P_1$ were merely degrading model behaviour at random, we would expect uniform confidence drops on CC and CW. Confidence drops are not uniform across outcome classes (Figure~\ref{fig:conf}). On the interventional rung, CW items show confidence drops two to five times larger than CC items in every model. Qwen2.5-14B is the most striking case, with a CW drop of $-0.135$ against a CC drop of $-0.025$. Models therefore produce wrong answers under $P_1$ with conscious uncertainty rather than with random noise.

\begin{table}[t]
\centering
\small
\setlength{\tabcolsep}{1pt}
\begin{tabular}{l ccc}
\toprule
\textbf{Model} & \textbf{Common $\Delta$} & \textbf{Nonsense $\Delta$} & \textbf{Engaged} \\
\midrule
Phi-3-mini & $+$6.2 & $+$0.6 & \cmark \\
Qwen2.5-7B & $+$9.8 & $+$0.0 & \xmark \\
Mistral-7B & $+$7.2 & $+$1.0 & \xmark \\
Llama-3.1-8B & $+$6.2 & $-$2.0 & \cmark \\
Qwen2.5-14B & $+$8.6 & $+$0.2 & \cmark \\
\midrule
\textbf{Mean (All models)} & \textbf{$+$7.6} & $\sim$0 & 3/5\\
\textbf{Mean (Engaged only)} & --- & \textbf{$-$0.4} & --- \\
\bottomrule
\end{tabular}
\caption{Within-benchmark falsifiability test on the interventional rung, comparing CLadder's commonsensical and nonsensical subsets.}
\label{tab:falsify}
\end{table}

\begin{table*}[t]
\centering
\small
\setlength{\tabcolsep}{9pt}
\begin{tabular}{l c ccc ccc ccc}
\toprule
& & \multicolumn{3}{c}{\textbf{CLadder}} & \multicolumn{3}{c}{\textbf{CRASS}} & \multicolumn{3}{c}{\textbf{e-CARE}} \\
\cmidrule(lr){3-5}\cmidrule(lr){6-8}\cmidrule(lr){9-11}
\textbf{Model} & \textbf{Size} & $\bm{P_0}$ & $\bm{P_1}$ & $\bm{\Delta}$ & $\bm{P_0}$ & $\bm{P_1}$ & $\bm{\Delta}$ & $\bm{P_0}$ & $\bm{P_1}$ & $\bm{\Delta}$ \\
\midrule
\multicolumn{11}{l}{\emph{Local (3.8B--14B)}} \\
Phi-3-mini       & 3.8B & 64.8 & 58.6 & $+$6.2 & 69.0 & 46.4 & $+$22.6 & 65.4 & 56.6 & $+$8.8 \\
Qwen2.5-7B$^\ddagger$ & 7B & 61.8 & 52.0 & $+$9.8 & 32.9 & 32.9 & $\phantom{+}$0.0 & 50.0 & 50.0 & $\phantom{+}$0.0 \\
Mistral-7B       & 7B   & 58.4 & 51.2 & $+$7.2 & 83.6 & 49.6 & $+$34.0 & 68.7 & 57.2 & $+$11.5 \\
Llama-3.1-8B     & 8B   & 59.8 & 53.6 & $+$6.2 & 73.4 & 42.7 & $+$30.7 & 67.6 & 57.5 & $+$10.1 \\
Qwen2.5-14B      & 14B  & 64.0 & 55.4 & $+$8.6 & 73.4 & 52.5 & $+$20.9 & 71.4 & 57.6 & $+$13.8 \\
\textbf{Local mean} & --- & \textbf{61.8} & \textbf{54.2} & \textbf{$+$7.6} & \textbf{74.8} & \textbf{47.8} & \textbf{$+$27.0} & \textbf{68.3} & \textbf{57.2} & \textbf{$+$11.1} \\
\midrule
\multicolumn{11}{l}{\emph{Frontier (70B and above)}} \\
Llama-3.3-70B    & 70B    & 70.6 & 60.8 & $+$9.8  & 94.2 & 71.9 & $+$22.3 & 82.3 & 62.8 & $+$19.5 \\
Qwen2.5-72B      & 72B    & 68.2 & 63.0 & $+$5.2  & 94.2 & 60.6 & $+$33.6 & 84.5 & 64.9 & $+$19.6 \\
Llama-4-Scout    & 109B$^\dagger$ & 63.8 & 57.0 & $+$6.8  & 82.8 & 55.8 & $+$27.0 & 70.8 & 58.3 & $+$12.5 \\
Llama-4-Maverick & 400B$^\dagger$ & 59.8 & 54.0 & $+$5.8  & 89.8 & 67.5 & $+$22.3 & 80.4 & 64.1 & $+$16.3 \\
Hermes-3-405B$^\star$ & 405B & 66.0 & 66.8 & $-$0.8 & 90.9 & 57.7 & $+$33.2 & 82.8 & 64.0 & $+$18.8 \\
DeepSeek-V3      & 671B$^\dagger$ & 65.0 & 54.8 & $+$10.2 & 93.4 & 61.3 & $+$32.1 & 81.9 & 65.4 & $+$16.5 \\
DeepSeek-V3.1    & 685B$^\dagger$ & 59.6 & 51.2 & $+$8.4  & 93.1 & 57.3 & $+$35.8 & 81.7 & 61.7 & $+$20.0 \\
DeepSeek-V3.2    & 685B$^\dagger$ & 60.6 & 50.8 & $+$9.8  & 90.5 & 52.9 & $+$37.6 & 82.3 & 61.8 & $+$20.5 \\
Qwen3-Max        & $\sim$1T$^\dagger$            & 63.8 & 58.8 & $+$5.0  & 95.5 & 72.7 & $+$22.8 & 82.8 & 64.4 & $+$18.4 \\
\textbf{Frontier mean} & --- & \textbf{64.2} & \textbf{57.5} & \textbf{$+$6.7} & \textbf{91.6} & \textbf{62.0} & \textbf{$+$29.6} & \textbf{81.1} & \textbf{63.0} & \textbf{$+$18.1} \\
\bottomrule
\end{tabular}
\caption{Cross-benchmark and cross-scale gap on CLadder (interventional rung), CRASS ($n=274$), and e-CARE ($n=1000$ stratified subsample). 
$^\ddagger$\,Qwen2.5-7B collapses on CRASS and e-CARE and is excluded from the engaged mean on those two benchmarks. $^\star$\,Hermes-3 uses non-standard SFT/DPO instruction tuning of Llama-3.1-405B-base. $^\dagger$\,Mixture-of-experts models reported at total parameter count.}
\label{tab:cross-benchmark}
\end{table*}

\subsection{Domain Analysis}
\label{sec:domain}

Per-question gaps are aggregated by CLadder's story domain, extracted from \texttt{story\_id} prefixes with manual merging of near-duplicates (\textit{smoke}/\textit{smoking} and \textit{encouagement}/\textit{encouragement}). The top five and bottom five domains ranked by cross-model mean gap are listed in Appendix Table~\ref{tab:domain}. 
The effect is sharply domain dependent. Domains with strong real-world causal priors show large positive gaps that are robust across models. Examples range from \emph{vaccine} at $+21.3$\,pp to \emph{smoking} at $+12.1$\,pp, with \emph{candle} and \emph{obesity} in between (full ranking in Appendix Table~\ref{tab:domain}). Domains that lack such priors show small or reversed effects, including \emph{water} at $-8.8$\,pp, \emph{nature}, and \emph{orange}. The observed heterogeneity rules out a uniform difficulty injection and points to memorized lexical priors as the mediator.

\subsection{Falsifiability Test}
\label{sec:falsify}

We re-ran the same five models on a 500-per-rung stratified sample of the nonsensical subset (Table~\ref{tab:falsify}). The cross-model mean interventional gap is $-0.04$\,pp across all five models, and $-0.4$\,pp restricted to the three models that remain engaged. This is a $19\!\times$ reduction from the commonsensical $+7.6$\,pp gap. The direction is no longer consistent across models, and every per-model gap satisfies $|\Delta| < 2.1$\,pp. 
A secondary finding is that on the nonsensical subset, Qwen2.5-7B emits \emph{no} on 100\% of interventional questions and Mistral-7B emits \emph{yes} on 92\% (Appendix Table~\ref{tab:engagement}). The same two models show much lower top-class shares on commonsensical CLadder, ranging from 71\% to 92\%, indicating greater output variation when lexical anchors are present. For these 7B-class models, lexical content of variable names functions not only as an accuracy shortcut but as a precondition for producing varied outputs on causal questions at all. The remaining three models stay engaged on nonsense yet score near chance.

\subsection{Cross-Benchmark Replication} \label{sec:cross}

\paragraph{Local and frontier set.} Per-model accuracy and gap are in Table~\ref{tab:cross-benchmark} and Figure~\ref{fig:cross-benchmark}. The cross-model mean gap on the engaged local models is $+27.0$ pp (CRASS) and $+11.1$ pp (e-CARE), both with every per-model gap positive. Across the nine frontier models the mean rises to $+29.6$ pp (CRASS) and $+18.0$ pp (e-CARE), again with every per-model gap positive. The lexical-fragility effect is therefore not a small-model artifact, and it persists across all frontier models. Frontier $P_0$ accuracy on CRASS reaches 90--96\%, so the larger frontier gaps reflect more absolute accuracy lost when lexical anchors are removed rather than a lower floor.

\paragraph{Item-level correlation.} An item-level analysis in Appendix~\ref{app:noun-density} finds a positive correlation between the number of replaced noun tokens and the per-item flip rate (Spearman $\rho = 0.23$ on CRASS, $\rho = 0.15$ on e-CARE), so denser noun content per item plausibly explains part of the magnitude difference between CRASS and CLadder.

\paragraph{Combined significance.} Across the benchmarks, all 13 model-benchmark pairs on the local engaged set show a positive gap. Adding the nine-model frontier set contributes another 27 pairs of which 26 are positive. The combined evidence is 39 of 40 positive pairs. A one-sided exact sign test against the null of zero direction gives $p \approx 3.7 \times 10^{-11}$.

\paragraph{Engagement collapse.} Qwen2.5-7B collapses to a single answer token (top-class share $1.00$) on CLadder nonsense, CRASS, and e-CARE, and Mistral-7B collapses on CLadder under perturbation. The other three local models and all nine frontier models remain engaged, indicating that engagement collapse is an intrinsic limitation of some 7B class models rather than a benchmark property.

\subsection{Prompting Mitigation} \label{sec:csp}

\paragraph{Prompting strategies} (see Table~\ref{tab:csp}). \textbf{Direct} is the minimal one-shot template that asks for a single \emph{Yes} or \emph{No} token and serves as the baseline used throughout the preceding analyses. \textbf{CausalCoT} \citep{jin2023cladder} is a four-step prompt originally validated on GPT-4 that asks the model to identify the causal graph, the query type, the necessary adjustment set, and then to compute the answer. \textbf{Scaffold} is a five-step structured scaffolding prompt we test as a candidate mitigation, asking the model to list the causal graph as edges, write the intervention as a $do(\cdot)$ specification, and provide a final answer in a strictly parseable form, with the exact template given in Appendix~\ref{app:prompts}. \textbf{Few-shot ICL} prepends five in-context demonstrations from the CLadder commonsensical interventional rung, disjoint from the evaluation pool and balanced three \emph{yes} to two \emph{no}, each presented in its original lexicalized form regardless of whether the test item is $P_0$ or $P_1$, with the same first-token logprob scoring as Direct.

\paragraph{Scaffold and Few-shot.} The scaffold reduces the cross-model mean gap from $+7.0$ pp under Direct to $+3.8$ pp, and Few-shot ICL further to $+1.4$ pp. The reduction is largely a $P_0$ regression on smaller models. Under the scaffold, Qwen2.5-14B drops 12.8 pp on $P_0$, Llama-3.1-8B 10.6 pp, and Phi-3-mini 10.0 pp. Under Few-shot, Qwen2.5-7B, Mistral-7B, and Phi-3-mini regress on $P_0$ by 6 to 11 pp; restricted to the four smaller models the Few-shot mean gap is still $+2.6$ pp. The only case where demonstrations enable an instruction-tuned LLM to solve $P_1$ items above its $P_0$ baseline is Qwen2.5-14B under Few-shot ($P_1 = 72.2$\% vs Direct $P_0 = 64.0$\%), matching the scale dependence in Figure~\ref{fig:scale}.

\paragraph{CausalCoT.} It produces no parseable yes/no answer in 14.6 to 70.1\% of cases across the five models, with mean $P_0$ accuracy of 30.9\%, well below chance. Our data indicate it does not scale-port to 7B--14B instruction-tuned models, in tension with its frequent use as a baseline in this regime.

\begin{table}[!t]
\centering
\small
\setlength{\tabcolsep}{1pt}
\begin{tabular}{l l ccc c}
\toprule
\textbf{Model} & \textbf{Method} & \textbf{$\bm{P_0}$ (\%)} & \textbf{$\bm{P_1}$ (\%)} & \textbf{$\bm{\Delta}$ (pp)} & \textbf{Invalid} \\
\midrule
Phi-3-mini & Direct & 62.0 & 56.2 & $+$5.8 & 0.2\% \\
 & CausalCoT & 44.4 & 33.0 & $+$11.4 & 35.2\% \\
 & Scaffold & 52.0 & 48.4 & $+$3.6 & 5.5\% \\
\addlinespace
Qwen2.5-7B & Direct & 63.2 & 52.6 & $+$10.6 & 0.0\% \\
 & CausalCoT & 16.8 & 13.8 & $+$3.0 & 70.1\% \\
 & Scaffold & 68.0 & 59.2 & $+$8.8 & 0.0\% \\
\addlinespace
Mistral-7B & Direct & 59.6 & 54.0 & $+$5.6 & 0.0\% \\
 & CausalCoT & 46.8 & 44.8 & $+$2.0 & 14.6\% \\
 & Scaffold & 61.2 & 57.0 & $+$4.2 & 0.2\% \\
\addlinespace
Llama-3.1-8B & Direct & 63.4 & 57.0 & $+$6.4 & 0.0\% \\
 & CausalCoT & 19.4 & 16.0 & $+$3.4 & 65.2\% \\
 & Scaffold & 52.8 & 50.8 & $+$2.0 & 11.3\% \\
\addlinespace
Qwen2.5-14B & Direct & 65.0 & 58.4 & $+$6.6 & 0.0\% \\
 & CausalCoT & 27.0 & 21.2 & $+$5.8 & 55.2\% \\
 & Scaffold & 52.2 & 51.8 & $+$0.4 & 0.1\% \\
\midrule
\textbf{Mean} & Direct       & 62.6 & 55.6 & \textbf{$+$7.0} & 0.0\% \\
\textbf{Mean} & CausalCoT    & 30.9 & 25.8 & \textbf{$+$5.1} & 48.1\% \\
\textbf{Mean} & Scaffold     & 57.2 & 53.4 & \textbf{$+$3.8} & 3.4\% \\
\bottomrule
\end{tabular}
\caption{Effect of prompting strategy on lexical fragility at the interventional rung.}
\label{tab:csp}
\end{table}

\section{Discussion} \label{sec:discussion}

\paragraph{Capability versus behaviour.} The lexical-mediation pattern is not an artifact of weak prompting at small scale. Scaling from 14B to 671B does not close the gap, and frontier mean gaps on CRASS and e-CARE exceed their local counterparts. The scaffold and few-shot ICL each reduce the CLadder gap on the local set, but on four of five models the reduction comes from $P_0$ regression rather than $P_1$ improvement. Only Qwen2.5-14B under few-shot improves $P_1$ above its $P_0$ baseline. Our claim is therefore restricted to standard zero-shot evaluation and is silent on what fine-tuning or richer elicitation might achieve.

\paragraph{Reasoning-tuned ablation.} As DeepSeek-R1 is reasoning-tuned rather than instruction-tuned, it lies outside the frontier set of Table~\ref{tab:cross-benchmark}. R1 shows positive gaps on all benchmarks ($+3.2$\,pp on CLadder, $+15.7$\,pp on CRASS, $+15.1$\,pp on e-CARE), uniformly smaller than the non-reasoning frontier means ($+6.7$, $+29.6$, $+18.0$\,pp respectively), suggesting that multi-step reasoning partially mitigates but does not eliminate lexical reliance. Invalid-output rates are elevated ($\sim$20\%) due to post-reasoning answer extraction. The gap direction is positive on all benchmarks regardless of whether invalid items are counted as wrong or excluded.

\section{Conclusion} \label{sec:conclusion}

We introduced \textbf{Caliper}, a controlled lexical perturbation that probes LLM causal reasoning while preserving the causal graph. Across nine instruction-tuned LLMs and three benchmarks, the gap localizes to ATE queries, collapses on a pseudoword subset, and resists three prompting interventions, indicating that reported accuracies are better understood as memorized lexical priors than direct evidence of causal inference.

\section*{Limitations} \label{sec:limitations}

We note three limitations. (1) Our model set spans open-weight instruction-tuned LLMs from 3.8B to 671B; closed-API proprietary models and models fine-tuned on causal-reasoning corpora are out of scope and may behave differently. (2) Mechanism analyses (query-type, domain, confidence) are conducted on CLadder only, since CRASS ($n=274$) has limited per-cell sample size, and we report only cross-model aggregates on CRASS and e-CARE. (3) The Caliper perturbation occasionally produces mildly ungrammatical phrases when a CLadder variable is a gerund (e.g.\ \emph{drinking}). The convergence of the gap on specific rungs, query types, and domains, together with its collapse on the nonsensical subset, rules out grammatical degradation as the dominant driver.

\section*{Ethics Statement}

This work probes the causal reasoning behaviour of publicly released instruction-tuned language models on three public benchmarks (CLadder, CRASS, e-CARE). No human subjects, personally identifiable information, or proprietary data are involved. Our findings identify a measurement limitation in standard zero-shot evaluation and are not a claim about the upper bound of what such models can do under fine-tuning or richer elicitation.

\bibliography{main}

\clearpage
\appendix

\section*{Appendix}
\addcontentsline{toc}{section}{Appendix}

\renewcommand{\thetable}{A\arabic{table}}
\renewcommand{\thefigure}{A\arabic{figure}}
\renewcommand{\thealgorithm}{A\arabic{algorithm}}
\setcounter{table}{0}
\setcounter{figure}{0}
\setcounter{algorithm}{0}

\section{Full Per-Model and Per-Query-Type Results} \label{app:full-qtype}

This appendix complements the aggregated results in Section~\ref{sec:exp} by providing the underlying per-model breakdowns. Each CLadder \texttt{query\_type} code label used in our tables and figures is mapped to its corresponding causal-inference term in Table~\ref{tab:querytype-glossary}. Per-model accuracy on $P_0$ and $P_1$, together with the lexical gap $\Delta$ for every (model, query-type) cell, is then reported in Table~\ref{tab:appendix-qtype}, from which the cross-model means and standard deviations summarised in Table~\ref{tab:querytype} are derived. Cells with $n < 5$ are omitted.

\begin{table}[h]
\centering
\small
\resizebox{\linewidth}{!}{
\setlength{\tabcolsep}{1pt}
\begin{tabular}{l c l}
\toprule
\textbf{Label} & \textbf{Rung} & \textbf{Full term} \\
\midrule
\texttt{marginal}       & Obs & marginal probability $P(Y)$ \\
\texttt{correlation}    & Obs & correlation $P(Y \mid X)$ \\
\texttt{exp\_away}      & Obs & explaining away \\
\texttt{ate}            & Int & average treatment effect (ATE) \\
\texttt{collider\_bias} & Int & collider bias \\
\texttt{backadj}        & Int & backdoor adjustment \\
\texttt{nde}            & CF  & natural direct effect (NDE) \\
\texttt{nie}            & CF  & natural indirect effect (NIE) \\
\texttt{ett}            & CF  & effect of treatment on the treated (ETT) \\
\texttt{det-cf}         & CF  & deterministic counterfactual \\
\bottomrule
\end{tabular}
}
\caption{Mapping between CLadder \texttt{query\_type} field values (used as column labels in tables and figures) and their full causal-inference terms.}
\label{tab:querytype-glossary}
\end{table}

\begin{table*}[!t]
\centering
\small
\begin{tabular}{lc c ccc c}
\toprule
\textbf{Model} & \textbf{Rung} & \textbf{Query Type} & \textbf{$\bm{n}$} & \textbf{$\bm{P_0}$ (\%)} & \textbf{$\bm{P_1}$ (\%)} & \textbf{$\bm{\Delta}$ (pp)} \\
\midrule
\multirow{10}{*}{Qwen2.5-14B}
 & Obs & marginal & 260 & 57.3 & 56.5 & $+$0.8 \\
 &     & correlation & 211 & 51.7 & 54.5 & $-$2.8 \\
 &     & exp\_away & 29 & 58.6 & 62.1 & $-$3.5 \\
 & Int & \textbf{ate} & 222 & 77.9 & 74.8 & \textbf{$+$3.1} \\
 &     & backadj & 252 & 53.6 & 39.7 & $+$13.9 \\
 &     & collider\_bias & 26 & 46.2 & 42.3 & $+$3.9 \\
 & CF  & det-counterfactual & 176 & 67.1 & 61.4 & $+$5.7 \\
 &     & ett & 171 & 57.3 & 55.0 & $+$2.3 \\
 &     & nde & 53 & 62.3 & 49.1 & $+$13.2 \\
 &     & nie & 100 & 52.0 & 44.0 & $+$8.0 \\
\midrule
\multirow{10}{*}{Qwen2.5-7B}
 & Obs & marginal & 260 & 48.1 & 46.5 & $+$1.6 \\
 &     & correlation & 211 & 49.8 & 49.3 & $+$0.5 \\
 &     & exp\_away & 29 & 48.3 & 65.5 & $-$17.2 \\
 & Int & \textbf{ate} & 222 & 68.9 & 55.4 & \textbf{$+$13.5} \\
 &     & backadj & 252 & 57.1 & 50.4 & $+$6.7 \\
 &     & collider\_bias & 26 & 46.2 & 38.5 & $+$7.7 \\
 & CF  & det-counterfactual & 176 & 61.9 & 58.5 & $+$3.4 \\
 &     & ett & 171 & 57.9 & 54.4 & $+$3.5 \\
 &     & nde & 53 & 47.2 & 54.7 & $-$7.5 \\
 &     & nie & 100 & 51.0 & 54.0 & $-$3.0 \\
\midrule
\multirow{10}{*}{Llama-3.1-8B}
 & Obs & marginal & 260 & 51.9 & 49.2 & $+$2.7 \\
 &     & correlation & 211 & 48.3 & 48.8 & $-$0.5 \\
 &     & exp\_away & 29 & 48.3 & 41.4 & $+$6.9 \\
 & Int & \textbf{ate} & 222 & 71.2 & 57.2 & \textbf{$+$14.0} \\
 &     & backadj & 252 & 47.6 & 48.0 & $-$0.4 \\
 &     & collider\_bias & 26 & 80.8 & 76.9 & $+$3.9 \\
 & CF  & det-counterfactual & 176 & 55.1 & 55.1 & $\phantom{+}$0.0 \\
 &     & ett & 171 & 49.1 & 48.0 & $+$1.1 \\
 &     & nde & 53 & 56.6 & 50.9 & $+$5.7 \\
 &     & nie & 100 & 58.0 & 47.0 & $+$11.0 \\
\midrule
\multirow{10}{*}{Mistral-7B-v0.3}
 & Obs & marginal & 260 & 47.3 & 49.2 & $-$1.9 \\
 &     & correlation & 211 & 54.0 & 50.2 & $+$3.8 \\
 &     & exp\_away & 29 & 51.7 & 51.7 & $\phantom{+}$0.0 \\
 & Int & \textbf{ate} & 222 & 69.8 & 55.4 & \textbf{$+$14.4} \\
 &     & backadj & 252 & 48.0 & 48.0 & $\phantom{+}$0.0 \\
 &     & collider\_bias & 26 & 61.5 & 46.2 & $+$15.4 \\
 & CF  & det-counterfactual & 176 & 58.0 & 55.7 & $+$2.3 \\
 &     & ett & 171 & 55.6 & 55.0 & $+$0.6 \\
 &     & nde & 53 & 54.7 & 50.9 & $+$3.8 \\
 &     & nie & 100 & 52.0 & 53.0 & $-$1.0 \\
\midrule
\multirow{10}{*}{Phi-3-mini}
 & Obs & marginal & 260 & 48.1 & 50.4 & $-$2.3 \\
 &     & correlation & 211 & 48.8 & 53.1 & $-$4.3 \\
 &     & exp\_away & 29 & 65.5 & 72.4 & $-$6.9 \\
 & Int & \textbf{ate} & 222 & 72.1 & 63.5 & \textbf{$+$8.6} \\
 &     & backadj & 252 & 59.1 & 55.2 & $+$4.0 \\
 &     & collider\_bias & 26 & 57.7 & 50.0 & $+$7.7 \\
 & CF  & det-counterfactual & 176 & 60.8 & 58.0 & $+$2.8 \\
 &     & ett & 171 & 50.9 & 48.0 & $+$2.9 \\
 &     & nde & 53 & 50.9 & 50.9 & $\phantom{+}$0.0 \\
 &     & nie & 100 & 48.0 & 47.0 & $+$1.0 \\
\bottomrule
\end{tabular}
\caption{Full per-model per-query-type accuracy and lexical perturbation gap for our evaluation. ATE rows are bolded to highlight the locus of the effect. Sample size $n$ refers to each cell's $P_0$ question count.}
\label{tab:appendix-qtype}
\end{table*}

\paragraph{Note on small sample cells.} Some observational sub-types such as \textit{exp\_away} have $n = 29$, which yields wider per-model confidence intervals. The cross-model aggregation reported in the main text reduces this noise.

\begin{table}[h]
\centering
\small
\setlength{\tabcolsep}{3.5pt}
\begin{tabular}{l cc cc}
\toprule
 & \multicolumn{2}{c}{\textbf{Commonsense}} & \multicolumn{2}{c}{\textbf{Nonsense}} \\
\cmidrule(lr){2-3}\cmidrule(lr){4-5}
\textbf{Model} & $\bm{P_0}$ & $\bm{P_1}$ & $\bm{P_0}$ & $\bm{P_1}$ \\
\midrule
Phi-3-mini & no\,65\% & no\,71\% & no\,82\% & no\,81\% \\
Qwen2.5-7B & no\,71\% & no\,72\% & no\,100\%$^\dagger$ & no\,100\%$^\dagger$ \\
Mistral-7B & yes\,81\% & yes\,92\%$^\dagger$ & yes\,92\%$^\dagger$ & yes\,92\%$^\dagger$ \\
Llama-3.1-8B & yes\,76\% & yes\,82\% & yes\,78\% & yes\,85\% \\
Qwen2.5-14B & no\,83\% & no\,64\% & no\,83\% & no\,58\% \\
\bottomrule
\end{tabular}
\caption{Top-class share of yes/no predictions on the interventional rung. $^\dagger$ marks $\geq 90\%$ single-class collapse. On commonsensical questions all 5 models remain engaged. On nonsensical questions Qwen2.5-7B and Mistral-7B collapse to majority predictions, providing direct evidence that lexical anchors are a prerequisite for inferential engagement.}
\label{tab:engagement}
\end{table}

\begin{table}[h]
\centering
\small
\begin{tabular}{l cc c}
\toprule
\textbf{Domain} & \textbf{Gap (pp)} & \textbf{$\bm{\sigma}$ (pp)} & \textbf{All positive?} \\
\midrule
\multicolumn{4}{l}{\textit{High-prior domains (top 5)}}\\
vaccine & $+$21.3 & 15.4 & \cmark \\
candle & $+$18.5 & 9.2 & \cmark \\
penguin & $+$18.5 & 20.4 & \xmark \\
encouragement & $+$17.1 & 8.6 & \cmark \\
obesity & $+$16.7 & 12.9 & \cmark \\
\midrule
\multicolumn{4}{l}{\textit{Neutral domains (bottom 5)}}\\
tax & $-$1.7 & 17.0 & \xmark \\
simpson & $-$4.2 & 9.7 & \xmark \\
orange & $-$4.4 & 5.4 & \xmark \\
nature & $-$5.5 & 4.5 & \xmark \\
water & $-$8.8 & 7.5 & \xmark \\
\bottomrule
\end{tabular}
\caption{Top 5 and bottom 5 story domains by interventional lexical gap (cross-model mean). The effect is concentrated in domains with strong real-world causal priors (vaccines, obesity, smoking) and is absent or reversed in neutral domains (water, nature, tax). $\sigma$ is across-model std-dev.}
\label{tab:domain}
\end{table}

\section{Qualitative Examples} \label{app:examples}

We provide three CW flips (correct on $P_0$, wrong on $P_1$) from Qwen2.5-14B on the interventional rung, selected by the largest confidence drop. Each example shows the original question ($P_0$), the lexically anonymized version ($P_1$), and the model's prediction with its confidence.

\paragraph{Example 1.} Gold answer is \emph{no}. $P_0$ prediction is \emph{no} with confidence $0.995$. $P_1$ prediction is \emph{yes} with confidence $0.527$. Variables in the original question are \emph{treatment}, \emph{unobserved}, \emph{drug}, \emph{freckles}.

\begin{quote}\small
\textbf{$P_0$ (original).}\ ``For patients not assigned the drug treatment, the probability of freckles is 30\%. For patients assigned the drug treatment, the probability of freckles is 29\%. For patients not assigned the drug treatment, the probability of taking of all assigned drugs is 18\%. For patients assigned the drug treatment, the probability of taking of all assigned drugs is 59\%. Will taking of all assigned drugs increase the chance of freckles?''
\end{quote}

\begin{quote}\small
\textbf{$P_1$ (lexically anonymized).}\ ``For patients not assigned the $X_1$ $X_2$, the probability of $X_3$ is 30\%. For patients assigned the $X_1$ $X_2$, the probability of $X_3$ is 29\%. For patients not assigned the $X_1$ $X_2$, the probability of taking of all assigned $X_1$ is 18\%. For patients assigned the $X_1$ $X_2$, the probability of taking of all assigned $X_1$ is 59\%. Will taking of all assigned $X_1$ increase the chance of $X_3$?''
\end{quote}

\paragraph{Example 2.} Gold answer is \emph{yes}. $P_0$ prediction is \emph{yes} with confidence $1.000$. $P_1$ prediction is \emph{no} with confidence $0.547$. Variables are \emph{drinking}, \emph{wife}, \emph{alarm}.

\begin{quote}\small
\textbf{$P_0$.}\ ``For people who do not drink coffee, the probability of ringing alarm is 42\%. For people who drink coffee, the probability of ringing alarm is 66\%. Will drinking coffee increase the chance of ringing alarm?''
\end{quote}

\begin{quote}\small
\textbf{$P_1$.}\ ``For people who do not drink coffee, the probability of ringing $X_1$ is 42\%. For people who drink coffee, the probability of ringing $X_1$ is 66\%. Will $X_2$ coffee increase the chance of ringing $X_1$?''
\end{quote}

\paragraph{Example 3.} Gold answer is \emph{no}. $P_0$ prediction is \emph{no} with confidence $0.988$. $P_1$ prediction is \emph{yes} with confidence $0.547$. Variables are \emph{treatment}, \emph{unobserved}, \emph{having}, \emph{cholesterol}.

\begin{quote}\small
\textbf{$P_0$.}\ ``For patients not assigned the drug treatment, the probability of low cholesterol is 71\%. For patients assigned the drug treatment, the probability of low cholesterol is 76\%. For patients not assigned the drug treatment, the probability of having a sister is 48\%. For patients assigned the drug treatment, the probability of having a sister is 8\%. Will having a sister increase the chance of low cholesterol?''
\end{quote}

\begin{quote}\small
\textbf{$P_1$.}\ ``For patients not assigned the drug $X_1$, the probability of low $X_2$ is 71\%. For patients assigned the drug $X_1$, the probability of low $X_2$ is 76\%. For patients not assigned the drug $X_1$, the probability of $X_3$ a sister is 48\%. For patients assigned the drug $X_1$, the probability of $X_3$ a sister is 8\%. Will $X_3$ a sister increase the chance of low $X_2$?''
\end{quote}

\paragraph{Observations.} In all three examples the $P_0$ prediction is high confidence (at least $0.99$) and correct. After lexical anonymization the confidence collapses to roughly $0.55$ and the answer flips. The pattern is consistent with the model exploiting lexical content (the implausibility of a causal link from \emph{drugs} to \emph{freckles}, or from \emph{sister} to \emph{cholesterol}) rather than performing structural inference over the supplied conditional probabilities. Examples~2 and~3 illustrate the grammatical artefact noted in the Limitations. When a variable name is a gerund (\emph{drinking}, \emph{having}), replacement produces mildly ungrammatical phrasing such as ``$X_2$ coffee''. The convergence of behavioural evidence across rung, query type, domain, and nonsense (\S\ref{sec:exp}) excludes grammar as the dominant mechanism.

\section{Prompt Templates} \label{app:prompts}

\paragraph{Direct prompt.} Used in Section \ref{sec:main} through Section \ref{sec:falsify}.
\begin{quote}\small\ttfamily
\{question\}\\[2pt]
Answer with a single token, Yes or No. After the token, optionally explain in one sentence.\\
Answer:
\end{quote}

\paragraph{CausalCoT prompt.} From \citet{jin2023cladder}. Used in Section \ref{sec:csp}.
\begin{quote}\small\ttfamily
\{question\}\\[2pt]
To answer this question, follow these steps.\\
1. Identify the causal structure (variables and direct relationships).\\
2. Determine the type of query (observational, interventional, or counterfactual).\\
3. Identify the necessary causal adjustment (e.g., backdoor variables).\\
4. Compute the answer based on the causal logic.\\
Then provide your final answer as Yes or No.
\end{quote}

\paragraph{Scaffold prompt.} Used in Section \ref{sec:csp}.
\begin{quote}\small\ttfamily
\{question\}\\[2pt]
To answer this rigorously, you MUST complete the following scaffold.\\[2pt]
[SCAFFOLD-1: CAUSAL\_GRAPH]\\
List ONLY the directed edges as a Python-style list of tuples.\\
Example: [(``smoking'', ``cancer''), (``smoking'', ``tar''),
(``tar'', ``cancer'')]\\
Edges =\\[2pt]
[SCAFFOLD-2: QUERY\_TYPE]\\
Choose exactly one. OBSERVATIONAL / INTERVENTIONAL / COUNTERFACTUAL.\\
Type =\\[2pt]
[SCAFFOLD-3: INTERVENTION\_SPEC] (skip if OBSERVATIONAL)\\
For INTERVENTIONAL or COUNTERFACTUAL, write do(VAR=value) explicitly.\\
do() =\\[2pt]
[SCAFFOLD-4: WORLDS] (mandatory for COUNTERFACTUAL only)\\
factual\_world = \{var: value, ...\}\\
counterfactual\_world = \{var: value, ...\}\\[2pt]
[SCAFFOLD-5: COMPUTATION]\\
Using only the scaffolds above, derive Yes or No in 1 to 3 sentences.\\
Reasoning:\\[2pt]
[FINAL]\\
Final Answer: Yes\\
or\\
Final Answer: No
\end{quote}

\paragraph{Generation hyperparameters.} For all three prompts we use greedy decoding with \texttt{do\_sample=False} and \texttt{pad\_token\_id = eos\_token\_id}. The \texttt{max\_new\_tokens} setting differs by prompt. For Direct we use 32, for CausalCoT 256, and for Scaffold 512.

\paragraph{Parser.} For Direct, we extract the first \emph{yes} or \emph{no} substring at the start of the generated text. For CausalCoT, we take the last \emph{yes} or \emph{no} substring in the output. For Scaffold, we apply a strict regex \texttt{Final\textbackslash s*Answer\textbackslash s*:?\textbackslash s*(Yes|No)} with a last yes or no fallback. Outputs that contain no \emph{yes} or \emph{no} token are counted as \emph{invalid}.

\section{Item-Level Noun-Density Analysis on CRASS and e-CARE} \label{app:noun-density}

The CRASS gap of $+27.0$\,pp is roughly $3.5\times$ the CLadder interventional-rung gap of $+7.6$\,pp (\S\ref{sec:cross}). A natural candidate explanation is that CRASS scenarios contain more replaceable noun tokens per item than CLadder questions, so each perturbed item loses more memorized lexical signal. We test this directly.

\paragraph{Setup.} For each item we record (i) the number of unique NOUN/PROPN tokens that the Caliper perturbation replaced, denoted $k$, and (ii) the per-item flip rate $f$, defined as the fraction of models that were correct on $P_0$ and wrong on $P_1$, restricted to items where at least two models were correct on $P_0$ (so $f$ is well-defined and not dominated by items the models never solve). We compute Pearson $r$ and Spearman $\rho$ between $k$ and $f$.

\paragraph{Results.} The correlations and the binned flip-rate pattern are reported in Table~\ref{tab:noun-density}. Both benchmarks show positive correlations, modest in magnitude. e-CARE shows a clean monotonic relationship across four bins, with flip rate rising from $20.9\%$ when 0--2 nouns are replaced to $36.9\%$ when 8 or more are replaced. CRASS shows a similar pattern but plateaus at the 3--4 bin, possibly because items with 5 or more replaceable nouns are rare in CRASS ($n=23$).

\begin{table}[h]
\centering
\small
\setlength{\tabcolsep}{4pt}
\begin{tabular}{l cccc}
\toprule
\textbf{Dataset} & \textbf{Pearson} $\bm{r}$ & \textbf{Spearman} $\bm{\rho}$ & $\bm{n}$ \textbf{items} \\
\midrule
CRASS  & $+0.179$ & $+0.231$ & 234 \\
e-CARE & $+0.154$ & $+0.148$ & 798 \\
\bottomrule
\end{tabular}\\[6pt]
\begin{tabular}{c c c c c}
\toprule
\multirow{2}{*}{\textbf{Bin ($\bm{k}$ nouns)}} & \multicolumn{2}{c}{\textbf{CRASS}} & \multicolumn{2}{c}{\textbf{e-CARE}} \\
\cmidrule(lr){2-3}\cmidrule(lr){4-5}
& \textbf{Flip Rate} & \textbf{$\bm{n}$ Items} & \textbf{Flip Rate} & \textbf{$\bm{n}$ Items} \\
\midrule
0--2 & 0.295 & 83 & 0.209 & 15 \\
3--4 & 0.454 & 126 & 0.218 & 157 \\
5--7 & 0.442 & 23 & 0.309 & 454 \\
8+ & --- & --- & 0.369 & 172 \\
\bottomrule
\end{tabular}
\caption{Item-level relationship between number of replaced noun/propn tokens ($k$) and per-item flip rate ($f$). Top: cross-item correlations. Bottom: $f$ binned by $k$. The 8+ bin contains no CRASS items: by construction of the CRASS test set, no item yields more than seven replaceable NOUN/PROPN tokens.}
\label{tab:noun-density}
\end{table}

\paragraph{Interpretation.} The positive correlations and the monotonic e-CARE binning support the hypothesis that more replaceable noun content per item leads to larger lexical fragility. The effect size is modest ($\rho \approx 0.15$--$0.23$), which means noun count alone does not account for the cross-benchmark difference in average gap. Other factors (MCQ choice count, topic familiarity, sentence-level structure) plausibly contribute. The finding is therefore best read as \emph{compatible with} a noun-density mechanism rather than as isolating it.

\begin{figure*}[t]
\centering
\includegraphics[width=0.8\linewidth]{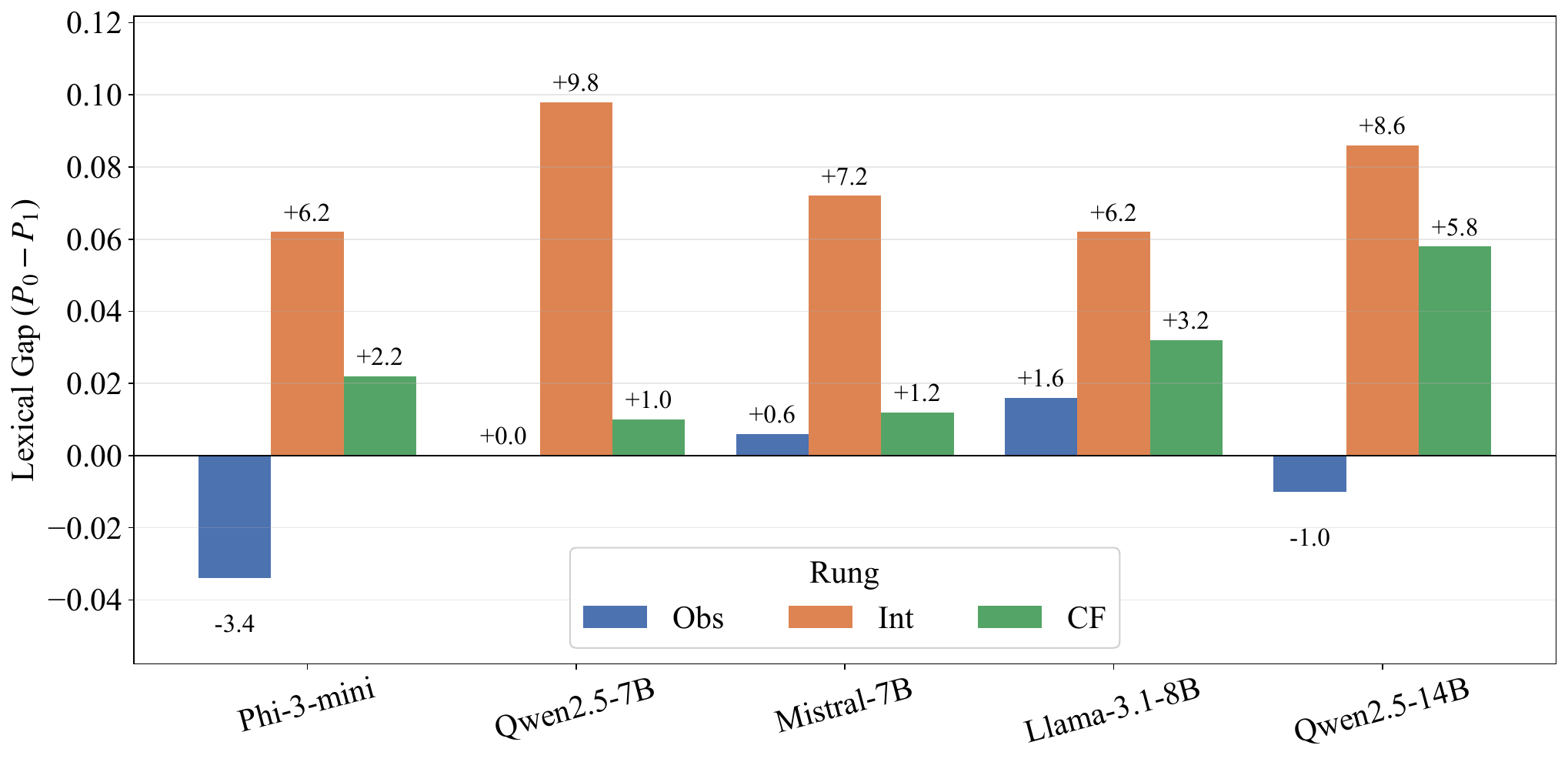}
\caption{Lexical perturbation gap ($P_0 - P_1$) by model and rung. The interventional gap (orange) is positive in all five models. The observational gap (blue) hovers around zero.}
\label{fig:rung}
\end{figure*}

\begin{figure*}[t]
\centering
\includegraphics[width=0.8\linewidth]{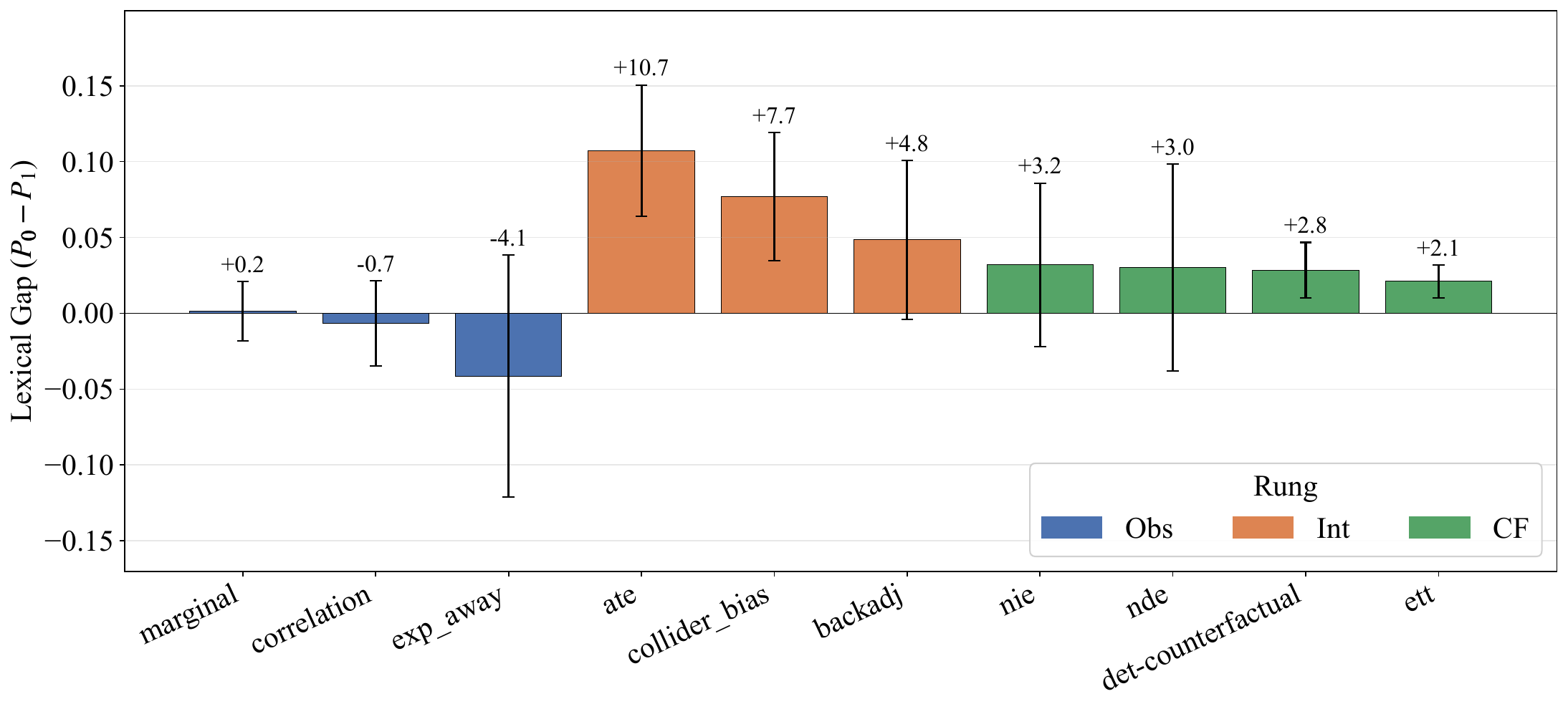}
\caption{Cross-model mean gap by query type with $\pm$ std-dev error bars ($n=5$ models). ATE shows the largest gap.}
\label{fig:qtype}
\end{figure*}

\begin{figure*}[t]
\centering
\includegraphics[width=\linewidth]{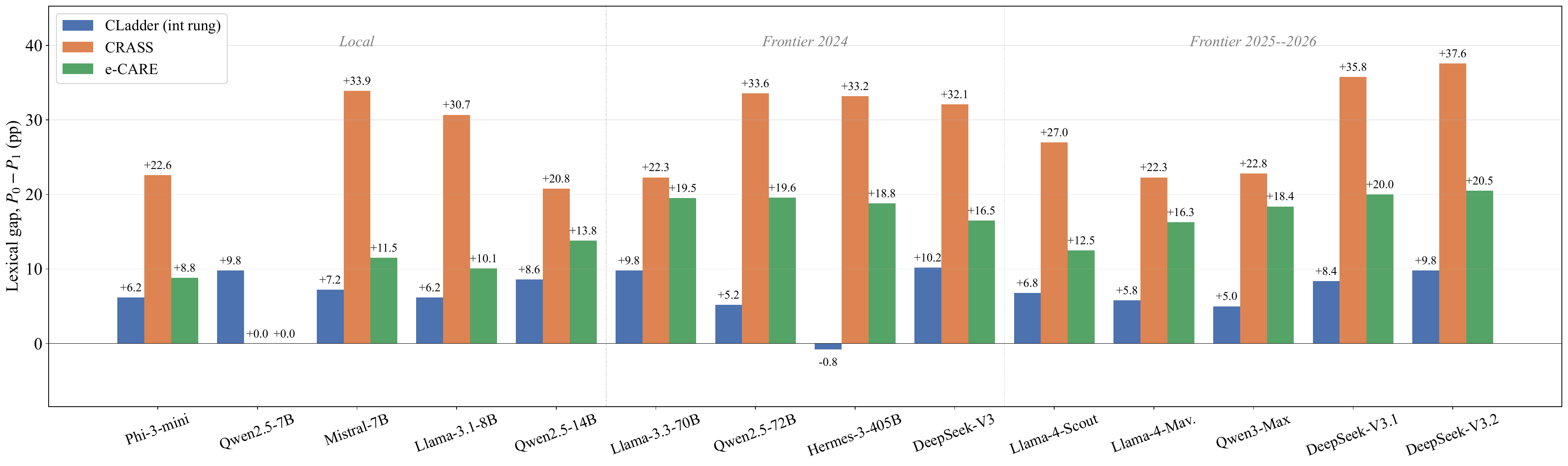}
\caption{Per-model lexical perturbation gap on three benchmarks across fourteen models (five local at 3.8B--14B and nine frontier from 70B to over 1T, separated by the vertical dotted line). Hatched bars mark engagement collapse (top-class share $\geq 0.90$), affecting Qwen2.5-7B on every benchmark in the local set, with no frontier model collapsing. The single direction reversal (Hermes-3-405B on CLadder, $-0.8$\,pp) is benchmark-specific, as the same model shows $+33.2$\,pp on CRASS and $+18.8$\,pp on e-CARE.}
\label{fig:cross-benchmark}
\end{figure*}

\end{document}